\documentclass[a4paper, onecolumn, oneside
    ]{IEEEtran}
\usepackage[utf8]{inputenc}
\usepackage{amsmath,amssymb,amsthm}
\usepackage{graphicx}
\usepackage[colorlinks=true]{hyperref}
\usepackage[capitalize]{cleveref}
\usepackage{booktabs}
\usepackage{array}
\usepackage{xfrac}
\usepackage{comment}
\usepackage[english]{babel}
\usepackage[
backend=biber,
style=ieee,
]{biblatex}
\AtEveryBibitem{
\clearfield{url}
    \clearfield{urldate}
    \clearfield{note}
    \clearfield{isbn}
    \clearfield{howpublished}
    \clearfield{Accessed}
    \clearfield{addendum}
    \clearfield{eprint}
    \clearfield{issn}
}
\usepackage{csquotes}
\usepackage{enumitem}

\usepackage{fancyhdr}
\title{Coupled Entropy: \\A Goldilocks Generalization for \\Complex Systems}
\author{Kenric P. Nelson\\
August 9, 2025}
\date{\today}

\newtheorem{lemma}{Lemma}
\newtheorem{definition}{Definition}
\newtheorem{proposition}{Proposition}

\begin{document}

\maketitle

\begin{abstract}
The coupled entropy is proven to correct a flaw in the derivation of the Tsallis entropy and thereby solidify the theoretical foundations for analyzing the uncertainty of complex systems. The Tsallis entropy originated from considering power probabilities $p_i^q$ in which \textit{q} independent, identically-distributed random variables share the same state. The maximum entropy distribution was derived to be a \textit{q}-exponential, which is a member of the shape ($\kappa$), scale ($\sigma$) distributions. Unfortunately, the $q$-exponential parameters were treated as though valid substitutes for the shape and scale. This flaw causes a misinterpretation of the generalized temperature and an imprecise derivation of the generalized entropy. The coupled entropy is derived from the generalized Pareto distribution (GPD) and the Student's t distribution, whose shape derives from nonlinear sources and scale derives from linear sources of uncertainty. The Tsallis entropy of the GPD converges to one as $\kappa\rightarrow\infty$, which makes it too cold. The normalized Tsallis entropy (NTE) introduces a nonlinear term multiplying the scale and the coupling, making it too hot. The coupled entropy provides perfect balance, ranging from $\ln \sigma$ for $\kappa=0$ to $\sigma$ as $\kappa\rightarrow\infty$. One could say, the coupled entropy allows scientists, engineers, and analysts to eat their porridge, confident that its measure of uncertainty reflects the mathematical physics of the scale of non-exponential distributions while minimizing the dependence on the shape or nonlinear coupling. Examples of complex systems design including a coupled variation inference algorithm are reviewed.
\end{abstract}

\section{Introduction}\label{Intro}
For several decades, an objective of complexity science, has been to define a generalization of entropy that separates the uncertainty due to the complexity from the uncertainty due to  non-complex components of the system. In this paper, I will show that statistical complexity can be measured as the nonlinearity creating a heavy-tailed (or compact-support) shape of the underlying distribution. And that the non-complex component of the system is the scale or generalized standard-deviation orginating from linear sources of uncertainty. From this foundation the coupled entropy is derived and shown to have uniquely important properties.

Entropy measures the disorder or average uncertainty of a system. One way to picture the concept of entropy is to understand that it is the average along the y, rather than the x axis of a distribution. The conceptual difficulty arises in that one must take the logarithm of the probabilities (or densities) in order for the arithmetic mean to be valid. Taking the inverse, $e^{-H}$, where H is the entropy, computes the average probability or density, and modifies the arithmetic mean into the geometric mean. For the simplest distribution, the exponential distribution, $f(x)=\frac{1}{\sigma}e^{-\frac{x}{\sigma}}$ , the entropy is $1+\ln\sigma$ and therefore the average density is $\frac{1}{e\sigma}=f(\sigma).$ That is the average density of the exponential distribution is the density at the scale.

This investigation of a proper generalization for complex systems, will start with the simplest distribution that includes a non-exponential shape, namely the generalized Pareto distribution (GPD). The background section will provide a more thorough description but to establish the goals of the paper, its sufficient to know that the entropy of the GPD is $1+\ln\sigma+\kappa$. That is in addition to the uncertainty due to the logarithm of the scale, there is a linear component due to the shape or nonlinear statistical coupling, $\kappa$, which can also be referred to as the complexity. The purpose of a generalized entropy is to minimize the influence of the complexity while retaining the dependence on the scale, i.e. the non-complex component.  

In Section \ref{Sec_Back}, I will derive the coupled exponential family.  Members of this family include the coupled exponential distribution (generalized Pareto), the coupled Weibull distribution, and the coupled Gaussian (Student's t). The parameter $q$ will be shown to decompose into more fundamental properties, $q=1+\frac{\alpha\kappa}{1+d\kappa}$, where $\alpha$ is the nonlinear parameter influencing the shape near the location, and $d$ is the dimensions of the distribution. In Section \ref{Sec_Entropy}, I derive the coupled entropy \cite{nelson_average_2017} function, which is related to the normalized Tsallis, and Tsallis entropies by

\begin{equation}
S_\kappa(\mathbf{p}) = \frac{1}{1+d\kappa}S_{\kappa}^{\text{NT}}(\mathbf{p})= \frac{1}{(1+d\kappa)\sum_jp_j^{1+\frac{\alpha\kappa}{1+d\kappa}}}S_{\kappa}^{\text{T}}(\mathbf{p}).
\end{equation}
The properties of these three generalized entropies are compared, showing that only the coupled entropy converges to the scale as the coupling (shape) goes to infinity. In Section \ref{Sec_Appl}, some properties of the coupled Boltzmann-Gibbs distribution are examined and a new approach to training robust variational inference algorithms \cite{nelson_variational_2025,blei_variational_2017} is reviewed. This new method uses the independent equals distribution to draw training samples. This allows for heavy-tailed (or compact-support) models, while drawing samples from distributions with faster (slower) decaying tails. The paper closes with concluding remarks and suggestions for future research. Prior to introducing the coupled entropy, I explain why the $q$-statistics formalism is inappropriate and must be replace by a framework grounded in the statistical definitions provided by Pareto and Gosset.

\section{Background}\label{Sec_Back}

\subsection{The priority of Pareto}

In the 1800s, Pareto \cite{arnold_pareto_2015} made important advances in the application of heavy-tailed statistics to the modeling of socioeconomic systems. The generalized Pareto distribution properly defines the independent role of the scale, $\sigma$ or spread of the distribution from the shape $\kappa$ of the asymptotic tail. If the survival function of the generalized Pareto distribution, is framed as a generalization of the exponential function, the algebra of the nonextensive statistical mechanics (NSM) \cite{borges_possible_2004,Tsallis2009a} can be set on a stronger foundation; $\exp_\kappa x \equiv (1 + \kappa x)_+^\frac{1}{\kappa}$, where $(a)_+\equiv \max(0,a)$. The location parameter is $\mu$.

\begin{definition}[Generalized Pareto Distribution]
\begin{align}\label{equ:survival}
    &\textit{Survival Function}\notag\\
    & S(x;\sigma,\kappa)\equiv\exp_\kappa^{-1}\left(\frac{x-\mu}{\sigma}\right)\equiv\left(1+\kappa\left(\frac{x-\mu}{\sigma}\right)\right)_+^{-\frac{1}{\kappa}} \\
    &(a)_+\equiv\max(0,a),\ x>\mu, \mu>0, \sigma>0, \kappa>-1\notag\\
    \notag\\
    &\textit{Probability Density Function}\notag\\
    &f_X(x;\sigma,\kappa)=\frac{-dS(x;\sigma,\kappa)}{dx}=
    \frac{1}{\sigma}\exp_\kappa^{-(1+\kappa)}\left(\frac{x-\mu}{\sigma}\right)\\
    &=\frac{1}{\sigma}\left(1+\kappa\left(\frac{x-\mu}{\sigma}\right)\right)_+^{-\frac{1+\kappa}{\kappa}}\notag
\end{align}
\end{definition}

The scale's independence from the shape of a distribution is established by two properties, the scaling of the random variable and the score function at the scale. Given $X\sim f_X(x;\sigma,\kappa)$, then
\begin{enumerate}
    \item $\frac{X}{\sigma}\sim f_X(x;1,\kappa)$
    \item $\sigma=-\left(\frac{d \ln f_X(x;\sigma,\kappa)}{dx}|_{x=\sigma}\right)^{-1}$
\end{enumerate}
These properties can be verified for the Generalized Pareto, Generalized Weibull, Student's t and many other shape-scale distributions.

\subsection{The problem with \textit{q}-statistics}
NSM began with the inquiry regarding the statistical mechanics of systems defined by an escort distribution \cite{tsallis_possible_1988},
\begin{equation}
    P^{(q)}\equiv\frac{p_i^q}{\sum_jp_j^q},
\end{equation}
leading to the Tsallis entropy and its thermodynamic formalism. The escort distribution extends to the real numbers the elementary property that $p_ip_i$ represents the probability of two independent random variables in the same state. The normalization makes $P^{(q)}$ a distribution. Thus, $q$ is the measure of the fractal number of independent random variables having the same state and is therefore referred to as the \textit{Independent Equals} \cite{nelson_independent_2022,al-najafi_independent_2024}. While the independent equals play an important role in specifying the sample distribution for moments, the central focus of complex systems needs to be the nonlinearity, which will be shown to be equal to the shape parameter. 

Unfortunately, to the author's knowledge there are no references in NSM literature stating the physical but secondary role of $q$ as a measure of the number of independent equals. Instead investigators using $q$-statistics have suggested that $q$ has a more significant but ill-defined purpose. Examples include:
\begin{itemize}
        \item ``Tsallis’ parameter q can be regarded as a measure of the degree of nonextensivity,” \cite{plastino_nonextensive_2000}
        \item ``the interpretation of the parameter q remains an open issue," \cite{Wilk2000}
        \item ``The entropic index q characterizes the degree of nonextensivity," \cite{abe_nonextensive_2001}
        \item ``a real number to be determined a priori from the microscopic dynamics," \cite{tsallis_nonextensive_2003}.
\end{itemize}
These descriptions are based on functions of $q$ rather than equality with a specific property.

The uninformative nature of the $q$-statistic parameters is shown through entropic analysis. Tsallis re-parameterized the generalized Pareto distribution (GPD) as the $q$-exponential distribution:
    \begin{align}
        f_X(x;\beta_q,q)\equiv
        \begin{cases}
            \beta_q (2-q) \left(1+(q-1)\beta x\right)_+^\frac{1}{1-q},& \text{if } q\neq1,\ (a)_+\equiv\max(0,a)\\
            \beta_1e^{-\beta_1 x},& q=1.
        \end{cases}
    \end{align}
The problem with this definition of a shape-scale distribution is made evident by comparing the expressions for the Boltzmann-Gibbs-Shannon (BGS) entropy:
\begin{align}
    &\begin{cases}
        H(f_X(x;\sigma,\kappa))=1+\kappa+\ln\sigma\\
        H(f_X(x;\beta_q,q))=1+\frac{q-1}{2-q}+\ln\left(\beta_q(2-q)\right)^{-1}.
    \end{cases}\\
    \\
    &\text{Matching terms gives the following relationships:}\\
&\kappa=\frac{q-1}{2-q},\ \qquad \sigma=\left(\beta_q(2-q)\right)^{-1}\\
&q=1+\frac{\kappa}{1+\kappa}, \qquad \beta_q=\frac{1+\kappa}{\sigma}.
\end{align}
First, the GPD definition is a clear simple description of the sources of uncertainty.  The evidence will show that the shape $\kappa$ depends only on the nonlinear or multiplicative sources of noise, while the scale $\sigma$ only depends on the linear or additive sources of noise. In contrast, the complicated entropy function for the $q$-exponential distribution is reflective of its uninformative parameters, which will make physical interpretation difficult, leading to otherwise easily avoidable errors. While $q$ only depends on $\kappa$ for the GPD we will see that this breaks down for multivariate distributions and for variables that have a $x^\alpha$ dependency. 

\begin{figure} \label{fig_scale}
    \centering
    \includegraphics[width=1\linewidth]{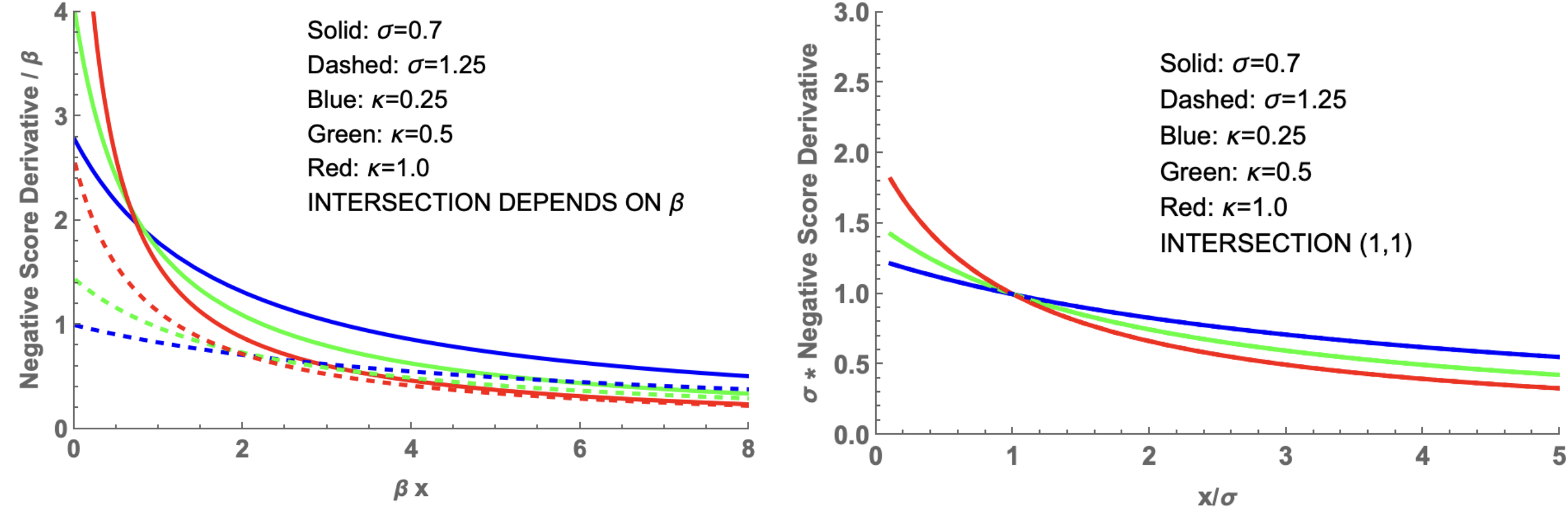}
    \caption{The q-exponential distribution (left) does not share a common intersection as the scale is changed. For the coupled exponential distribution (GPD) (right) normalization by the scale assures that the full family instersects at (1,1).}
    \label{fig:scale}
\end{figure}

Worse, $\beta_q$ is the ratio of $1+\kappa$ and $\sigma$, clarifying that it is not the scale of the distribution. Inappropriately, the NSM literature has defined a generalized temperature to be $T_q\equiv(k_B \beta_q)^{-1}$ \cite{abe_temperature_2006, tsallis_introduction_2004}, where $k_B$ is the Boltzmann constant. The consequence is the physical absurdity that the independent linear and nonlinear sources of temperature can both go to infinity while $T_q$ remains finite. Instead, a generalized temperature must be defined based on the scale of the distribution so that it is independent of the nonlinearity, $T_\kappa\equiv\frac{\sigma}{k_B}$. Figure \ref{fig:scale} shows that $\beta_q$ is not invariant to the shape of the distribution, while the GPD does define the scale such that it is invariant to the shape. A related approach to clarifying the actual scale of the distribution is the negative score function:
\begin{align}
    -\frac{d\ln(\text{Exp}_\kappa(\sigma))}{dx}\Big|_{x=\sigma}=\sigma^{-1}=\beta_q(1-q).
\end{align}
Thus the inverse-scale of the $q$-exponential distribution is not $\beta_q$.

Next, the GPD will be encompassed within the coupled exponential family of distributions.

\subsection{The Coupled Exponential Family}

The coupled exponential family \cite{nelson_definition_2015} provides a unification of many heavy-tailed (and compact-support) distributions while preserving the priority of properly defining the shape and scale of the distributions, and thereby establishing the proper foundation for the physical analysis of complex systems. Use of a generalized exponential function, requires starting with the survival function, otherwise the measure of dimensions $d$, which impacts the probability distribution function, is not properly separated. Next, it is important to distinguish between the shape of the distribution near the location $(\alpha: x\rightarrow\mu)$ and the asymptotic shape of the distribution $(\kappa: x\rightarrow\infty)$.   To do so, the exponent of the survival function is $-\sfrac{1}{\alpha\kappa}$, so that $\alpha$'s influence diminishes as $x\rightarrow\infty$, and the power-law tail decay is determined by $\kappa$ alone. Likewise, for finite $\kappa$, as $x\rightarrow 0$ the survival function approximates a stretched exponential with $\alpha$ determining the curvature. In Section \ref{Sec_Entropy}, this definition will be shown to be unique in properly defining the scale and shape of the distribution.

\begin{definition}[Coupled Survival Function]
For location $\mu$, scale $\sigma$, and shape $\alpha$:
\begin{equation}
S_\kappa(x;\mu,\sigma,\alpha)
\equiv \left(1 + \kappa\left(\frac{x-\mu}{\sigma}\right)^\alpha\right)^{-\frac{1}{\alpha\kappa}}
\equiv\exp_\kappa^{-\sfrac{1}{\alpha}}\left(\frac{x-\mu}{\sigma}\right)^\alpha\ 
\end{equation}
\end{definition}

The derivative of the cumulative distribution, $1-S_\kappa(x)$, gives the coupled probability density functions. Shown below are the normalizations for the heavy-tailed domain, $\kappa>0$. For $-1<\kappa<0$, the structure is the same with a modification to the normalization, given the cutoff at $x=\mu+\frac{\sigma}{\kappa^\alpha}$.

\begin{proposition}[Coupled Density Functions]
The general coupled density function is :
\begin{align}
p_\kappa(x) &= \frac{1}{\sigma}\left(\frac{x-\mu}{\sigma}\right)^{\alpha-1}\left(1+\kappa\left(\frac{x-\mu}{\sigma}\right)^\alpha\right)^{-\left(\frac{1}{\alpha\kappa}+1\right)}
\equiv \frac{1}{\sigma}\left(\frac{x-\mu}{\sigma}\right)^{\alpha-1}\exp_\kappa^{-\frac{1+\alpha\kappa}{\alpha}}\left(\frac{x-\mu}{\sigma}\right)^\alpha.
\end{align}
The PDFs for $\alpha=1,2$ are the generalized Pareto and generalized Weibull distributions:
\begin{align}
p_\kappa^{\text{exp}}(x) &\equiv \frac{1}{\sigma}\left(1+\kappa\frac{x-\mu}{\sigma}\right)^{-\frac{1+\kappa}{\kappa}}\label{equ_expdist} 
\equiv \frac{1}{\sigma}\exp_\kappa^{-(1+\kappa)}\left(\frac{x-\mu}{\sigma}\right).\\
p_\kappa^{\text{Weib}}(x) &\equiv \frac{x-\mu}{\sigma^2}\left(1+\kappa\left(\frac{x-\mu}{\sigma}\right)^2\right)^{-\left(\frac{1}{2\kappa}+1\right)}
\equiv \frac{1}{\sigma}\left(\frac{x-\mu}{\sigma}\right)\exp_\kappa^{-\frac{1+2\kappa}{2}}\left(\frac{x-\mu}{\sigma}\right)^2.
\end{align}
If the $\left(\frac{x-\mu}{\sigma}\right)^{\alpha-1}$ term is dropped from the pdf and the exponent is adjusted, we have definition for the coupled stretched exponential distribution 
\begin{align}
p_\kappa^{\text{Stretch}}(x) &\equiv \frac{1}{Z}\left(1+\kappa\left(\frac{x-\mu}{\sigma}\right)^\alpha\right)^{-\frac{1+\kappa}{\alpha\kappa}}
\equiv \frac{1}{Z}\exp_\kappa^{-\frac{1+\kappa}{\alpha}}\left(\frac{x-\mu}{\sigma}\right)^\alpha.
\end{align}
For $\alpha=2$, the distribution is the coupled Gaussian 
\begin{align}
p_\kappa^{\text{Gauss}}(x) &\equiv \frac{1}{Z}\left(1+\kappa\left(\frac{x-\mu}{\sigma}\right)^2\right)^{-\frac{1+\kappa}{2\kappa}}
\equiv \frac{1}{Z}\exp_\kappa^{-\frac{1+\kappa}{2}}\left(\frac{x-\mu}{\sigma}\right)^2.
\end{align}
where $Z=\sigma\sqrt{\pi/\kappa}\frac{\Gamma(1/(2\kappa))}{\Gamma((1+\kappa)/(2\kappa))}$. 
\end{proposition}
The coupled Gaussian is equal to the Student's t distribution with $\kappa=1/\nu$, where $\nu$ is the degree of freedom. Its survival function is a hypergeometric function. If these distributions are extended to multivariates, the term $1+\kappa$ in the exponent becomes $1+d\kappa$, throughout. For simplicity, this introduction to the coupled entropy will use the multivariate form when helpful without a full derivation.

In the heavy-tailed domain, $\kappa>0,$ the moments, $\mu_m=\int_Xx^mp(x)dx$ of the coupled exponential distributions are either undefined or divergent for $\kappa\geq\sfrac{1}{m}.$ Nevertheless, the following independent equals moments are finite for all $\kappa$ \cite{nelson_independent_2022,al-najafi_independent_2024}, 
\begin{equation}
    \mu_m^{(1+\frac{m\kappa}{1+\kappa})}=\int_Xx^m P^{\left(1+\frac{m\kappa}{1+\kappa}\right)}(x)dx;\ \ P^{\left(q\right)}(x)\equiv\frac{p^q(x)}{\int_Xp^q(x)dx}.
\end{equation}
The independent equals moments are not dependent on $\alpha$. For $p_\kappa^{\text{exp}}(x)$, $\mu_1^{(1+\frac{\kappa}{1+\kappa})}=\sigma$, and for $p_\kappa^{\text{Gauss}}(x)$, $\mu_1^{(1+\frac{\kappa}{1+\kappa})}=\mu$ and $\mu_2^{(1+\frac{2\kappa}{1+\kappa})}=\sigma^2$.

\subsection{Additive and Multiplicative Noise Sources}
A crucial aspect of the shape-scale distributions is their separation of additive (linear) and multiplicative (nonlinear) noise sources. The term nonlinear statistical coupling \cite{nelson_nonlinear_2010} is used for both this field of inquiry and the shape parameter in part from the proof linking the shape and the nonlinear source. There are several derivations of this linkage including, \textcite{beck_superstatistics_2003, nelson_average_2017} model of superstatistics fluctuations and \textcite{hahn_q-gaussians_2010} model of exchangeability of random variables. Here I show the lemmas based on multiplicative noise processes. The proof is in Appendix \ref{App_Noise}.
\begin{lemma}[Multiplicative Process with Coupled Gaussian Limit]
Let a stochastic process $X_t$ be defined by the Stratonovich differential equation,
\begin{equation*}
    dX_t=\underbrace{f(X_t)}_{\text{Drift}}dt+\underbrace{A \circ dW_t^{(a)}}_{\text{Additive Noise}}+\underbrace{g(X_t)M \circ dW_t^{(m)}}_{\text{Multiplicative Noise}}
\end{equation*}
where $dW^{(a)}_t$ and $W^{(m)}_t$ are independent Wiener processes which define the additive (a) and multiplicative (m) noise, and A and M are the amplitudes of each noise source.  Let the drift, $J(x)=f(X_t)$, be related to the diffusion, $D(x)=\frac{1}{2}(A^2 + M^2 g^2(x))$ by a restorative potential, $V(x)$, in which $f(x)=-\tau g(x)g'(x)=-V'(x)$ and $\tau$ is a time-domain scaling. Then the probability density $p_X(x,t)$ for this system has a coupled Gaussian limit distribution of $p_X(x)=\lim_{t\to\infty}p_X(x,t)\propto \exp^{-\frac{1+\kappa}{2}}_\kappa \left(\frac{g^2(x)}{\sigma^2}\right)$, with $\sigma^2=\frac{A^2}{2\tau}$ and $\kappa =\frac{D'(x)}{V'(x)}=\frac{M^2}{2\tau}$.
\end{lemma}

Having clarified the significance of the shape and scale in separating the nonlinear and linear sources of uncertainty, I will now define the coupled entropy and its properties.

\section{Coupled Entropy Foundations}\label{Sec_Entropy}

Proper definition of the shape-scale distributions motivates examination of the generalized entropy for nonextensive statistical mechanics. While the inverse of the coupled exponential function, gives the coupled logarithm $\ln_\kappa x=\frac{1}{\kappa}\left(x^\kappa-1\right)$, the coupled pdfs require a different power term as a result of the derivatives from the cdf.  Inverting the coupled stretched exponential distribution and considering the multivariate form without the location and scale, gives
\begin{align}
    y^\alpha =\left(1+\kappa\left(\boldsymbol{X^{-1}X}\right)^\frac{\alpha}{2}\right)^{-(1+d\kappa)/\kappa}\rightarrow\left(\boldsymbol{X^{-1}X}\right)^\frac{\alpha}{2}=\frac{1}{\kappa}\left((y^\alpha)^{-\frac{\kappa}{1+d\kappa}}-1\right).
\end{align}
The first definition of the coupled entropy, treats $\alpha$ as part of the arguments and thereby, provides a generalization of the entropy focused on the coupling parameter with the dimensional parameter providing the proper scaling for multivariate analysis. A second definition of the coupled entropy includes $\alpha$ by raising the argument to the power $\alpha$ and take the root $1/\alpha$ of the result. Since the root complicates computations, a third definition, removes the root but adds a multiplicative term of $\sfrac{1}{\alpha}$ to assure convergence to the entropy of the Gaussian for $\kappa\rightarrow0$.

The coupled entropy averages the coupled logarithm of the probabilities over the independent equals distribution. This is consistent with taking generalized moments. This is similar to the normalized Tsallis entropy; however, the coupled logarithm term has a distinction between its multiplicative term $\sfrac{1}{\kappa}$ and its exponent $-\frac{\kappa}{1+d\kappa}$. This has the effect of the coupled entropy being divided by $1+d\kappa$ relative to the normalized Tsallis entropy. I'll show in the next section that this provides a unique result, that balances issues with Tsallis and normalized Tsallis entropies, regarding a generalized entropy for the coupled exponential (generalized Pareto) distribution. 

\subsection{Core Definitions}

\begin{definition}[Coupled Entropy]
\begin{equation}\label{equ_CEI}
\text{I.  } H_\kappa(\mathbf{p}) = \sum_{i_1...i_j..i_d} P_{i_j}^{\left(1+\frac{\kappa}{1+d\kappa}\right)}\ln_\kappa\left(p_{i_j}\right)^{-\frac{1}{1+d\kappa}}=\frac{1}{\kappa}  \left(-1+\left(\sum_{i_1...i_j..i_d} p_{i_j}^{1+\frac{\kappa}{1+d\kappa}}\right)^{-1}\right)
\end{equation}
\begin{equation}
\text{II.  } H_{\alpha,\kappa}(\mathbf{p}) = \sum_{i_1...i_j..i_d} P_{i_j}^{\left(1+\frac{\kappa}{1+d\kappa}\right)}\left(\ln_\kappa\left(p_{i_j}\right)^{-\frac{\alpha}{1+d\kappa}}\right)^\frac{1}{\alpha}
\end{equation}
\begin{align}
\text{III.   } H_{\alpha,\kappa}(\mathbf{p}) &= \sum_{i_1...i_j..i_d} P_{i_j}^{\left(1+\frac{\alpha\kappa}{1+d\kappa}\right)}\left(\frac{1}{\alpha}\ln_\kappa\left(p_{i_j}\right)^{-\frac{\alpha}{1+d\kappa}}\right)\\
&= \sum_{i_1...i_j..i_d} P_{i_j}^{\left(1+\frac{\alpha\kappa}{1+d\kappa}\right)}\ln_{\alpha\kappa}\left(p_{i_j}\right)^{-\frac{1}{1+d\kappa}}
\end{align}
where $P_{i_j}^{\left(1+\frac{\kappa}{1+d\kappa}\right)}= \frac{p_{i_j}^{1+\frac{\kappa}{1+d\kappa}}}{\sum\limits_{k_1...k_l...k_d} p_{k_l}^{1+\frac{\kappa}{1+d\kappa}}}$ and $\sum_{i_1...i_j..i_d}$ is the summation over multiple dimensions.
\end{definition}
The information entropy symbol $H$ is used here, though the methods extend to statistical mechanics with the Boltzmann factor. The comparison with Tsallis entropies focuses on Type I. The application of Type III to machine learning is discussed in Section \ref{Sec_Appl}. Discussion of Type II is deferred for further research. For simplicity, the remainder of the paper will only show one-dimensional indices; however, the other dimensions are implied when the dimensional variable is included.

The distinction between the exponent applied to the probability, $-\frac{\alpha\kappa}{1+d\kappa}$ and the normalization $\frac{1}{\alpha\kappa}$ for the coupled logarithm has a number of advantages. For the Type III coupled entropy, the multiplication of the two nonlinear terms defines the degree of non-additivity for independent systems:
\begin{equation}
H_{\alpha,\kappa}(A+B) = H_{\alpha,\kappa}(A) + H_{\alpha,\kappa}(B) + \alpha\kappa H_{\alpha,\kappa}(A)H_{\alpha,\kappa}(B).
\end{equation}
The relationship can be notated using the coupled sum as $H_{\alpha,\kappa}(A) \oplus_{\alpha\kappa} H_{\alpha,\kappa}(B)$. Thus the two nonlinear parameters multiply to determine the magnitude of the non-additivity. In contrast, the nonextensive character of the coupled exponential is defined by the exponent, $r=\frac{\alpha\kappa}{1+d\kappa}$, referred to as the information relative risk aversion \cite{nelson_reduced_2020}, given its connection to the concept of relative risk aversion. For the heavy-tailed domain, $0<\kappa<\infty$, the relative risk aversion has a range of $0<r<\frac{\alpha}{d}$. Given a system which the effective number of equiprobable states, $W$, scales with a power of  the number of the states, $W(N)\sim N^\rho$, a risk exponent of $r=\sfrac{1}{\rho}$ creates extensive growth in the coupled entropy since:
\begin{align}
    H_{\kappa,\alpha}(N)=\ln_{\alpha\kappa}N^{\frac{\rho}{1+d\kappa}}=\frac{N^{\frac{\rho\alpha\kappa}{1+d\kappa}}-1}{\alpha\kappa}.
\end{align}
\subsection{Coupled Cross-Entropy and Divergence}

The coupled cross-entropy follows naturally as 
\begin{equation}
    H_\kappa (\textbf{p}\|\textbf{q})=\sum_i P_i^{\left(1+\frac{\kappa}{1+d\kappa}\right)}\ln_\kappa\left(q_i\right)^{-\frac{1}{1+d\kappa}}
\end{equation}
Two definitions of the coupled divergence are possible depending on whether the coupled sum is utilized, 
\begin{definition}[Coupled Divergence]
    \begin{align}
        \text{I.  }&D_\kappa(\textbf{p}\|\textbf{q})=H_\kappa (\textbf{p})-H_\kappa (\textbf{p}\|\textbf{q})=\sum_i P_i^{1+\frac{\kappa}{1+d\kappa}}\left(\ln_\kappa p_i^{-\frac{1}{1+d\kappa}}-\ln_\kappa
        q_i^{-\frac{1}{1+d\kappa}}\right)\\
        \text{II.  }&D_\kappa(\textbf{p}\|\textbf{q})=H_\kappa (\textbf{p})\ominus_\kappa H_\kappa (\textbf{p}\|\textbf{q})=\sum_i P_i^{1+\frac{\kappa}{1+d\kappa}}\ln_\kappa \left(\frac{p_i}{q_i}\right)^{-\frac{1}{1+d\kappa}}
    \end{align}
\end{definition}
A promising area of research will be to apply information geometry to evaluating the relative merit of these two definitions for the coupled divergence.

\subsection{Comparison of Generalized Entropy Properties}

A comparison of the coupled entropy with the Tsallis and normalized Tsallis entropies for the coupled exponential (generalized Pareto) distribution, provides a clear illustration of a) why the normalized Tsallis entropy has a stability issue, b) why the Tsallis entropy is stable but is insufficient in measuring the uncertainty due to the scale of the distribution, and c) how the coupled entropy can provide a unique generalization of entropy for measuring the uncertainty of heavy-tailed distributions. For simplicity, I'll focus on the one-dimensional distribution.

Given X distributed as a coupled exponential distribution:
\begin{equation}
    X\sim \text{Exp}_\kappa{\frac{1}{\sigma}}:\ p_\kappa^{\text{exp}}(x;\sigma)= \frac{1}{\sigma}\left(1+\kappa\frac{x}{\sigma}\right)^{-(1+\kappa)/\kappa}
\end{equation}
I. Shannon Entropy \\
The Shannon Entropy measures the uncertainty of X as one plus the logarithm of the scale plus the coupling.
\begin{align}
    H(p_\kappa^{\text{exp}}(x;\sigma))=\int_0^\infty \frac{1}{\sigma}\left(1+\kappa\frac{x-\mu}{\sigma}\right)^{-(1+\kappa)/\kappa} &\ln\left(\frac{1}{\sigma}\left(1+\kappa\frac{x-\mu}{\sigma}\right)^{(1+\kappa)/\kappa} \right)\\
    =1+\ln\sigma+\kappa\\
\end{align}
 II. Tsallis Entropy with $q=1+\frac{\kappa}{1+\kappa}=\frac{1+2\kappa}{1+\kappa}$ \\
 The Tsallis entropy subtracts from one coupled logarithm of the inverse of the scale. As such, the Tsallis entropy is not directly measuring the generalized logarithm of the uncertainty due to the scale. Further a divisive factor $1+\kappa$ diminishes the influence of the scale measurement as the complexity $\kappa$ increases.
 \begin{align}
     H_\kappa^T(p_\kappa^{\text{exp}}(x;\sigma))=\int_0^\infty 
     \frac{1}{\sigma}\left(1+\kappa\frac{x}{\sigma}\right)^{-\frac{1+\kappa}{\kappa}\frac{1+2\kappa}{1+\kappa}}
     &(1+\kappa)\ln_\kappa\left(\frac{1}{\sigma}\left(1+\kappa\frac{x}{\sigma}\right)^{\frac{1+\kappa}{\kappa}} \right)\\
    =1-\frac{1}{1+\kappa}\ln_\frac{\kappa}{1+\kappa} &{\sigma^{-1}}
 \end{align}
 III. Normalized Tsallis Entropy\\
 The Normalized Tsallis entropy of the coupled exponential distribution includes a term that multiples the coupling with the coupled logarithm of the scale causing it to rapidly increase to infinity.

 The normalization factor is equal to:
 \begin{align}
     \int_0^\infty 
     \left(\frac{1}{\sigma}\right)^\frac{1+2\kappa}{1+\kappa}\left(1+\kappa\frac{x}{\sigma}\right)^{-\frac{1+2\kappa}{\kappa}}
     =\frac{\sigma^{-\frac{\kappa}{1+\kappa}}}{1+\kappa}
 \end{align}
 The Normalized Tsallis entropy of the coupled exponential distribution is:
 \begin{align}
         H_\kappa^{NT}(p_\kappa^{\text{exp}}(x;\sigma))=\frac{1+\kappa}{\sigma^{-\frac{\kappa}{1+\kappa}}}\int_0^\infty 
     \left(\frac{1}{\sigma}\right)^\frac{1+2\kappa}{1+\kappa}\left(1+\kappa\frac{x}{\sigma}\right)^{-\frac{1+2\kappa}{\kappa}}
     &(1+\kappa)\ln_\kappa\left(\frac{1}{\sigma}\left(1+\kappa\frac{x}{\sigma}\right)^{-\frac{1+\kappa}{\kappa}} \right)^\frac{1+2\kappa}{1+\kappa}\\
    =1+(1+\kappa)\ln_\frac{\kappa}{1+\kappa} &{\sigma}+\kappa
 \end{align}
 III. Coupled Entropy\\
 The coupled entropy uniquely satisfies the purpose of a generalized entropy for complex systems regarding a measure of uncertainty for the coupled exponential distribution. The uncertainty is measured as one plus the coupled logarithm of the scale. The coupling value of the logarithm is $r=\frac{\kappa}{1+\kappa}$, which is the information relative risk aversion. Thus as the complexity goes to infinity $\kappa\rightarrow\infty$, the relative risk aversion goes to one, $r\rightarrow 1$. Thereby, the range of coupled logarithms utilized is tempered between $ H_0(p_0^{\text{exp}}(x;\sigma))=1+\ln\sigma$ and $ H_{\kappa\rightarrow\infty}(p_{\kappa\rightarrow\infty}^{\text{exp}}(x;\sigma))=1+\ln_1\sigma=\sigma$.
 \begin{align}
         H_\kappa(p_\kappa^{\text{exp}}(x;\sigma))&=\frac{1+\kappa}{\sigma^{-\frac{\kappa}{1+\kappa}}}\int_0^\infty 
     \frac{1}{\sigma}\left(1+\kappa\frac{x}{\sigma}\right)^{-\frac{1+2\kappa}{1+\kappa}}
     \ln_\kappa\left(\frac{1}{\sigma}\left(1+\kappa\frac{x}{\sigma}\right)^{\frac{1+\kappa}{\kappa}} \right)\\
 \end{align}
 {\Large\begin{equation*}
     \boxed{
     H_\kappa(p_\kappa^{\text{exp}}(x;\sigma))=1+\ln_\frac{\kappa}{1+\kappa} {\sigma}
 }
 \end{equation*}}

\begin{figure}
    \centering
    \includegraphics[width=1\linewidth]{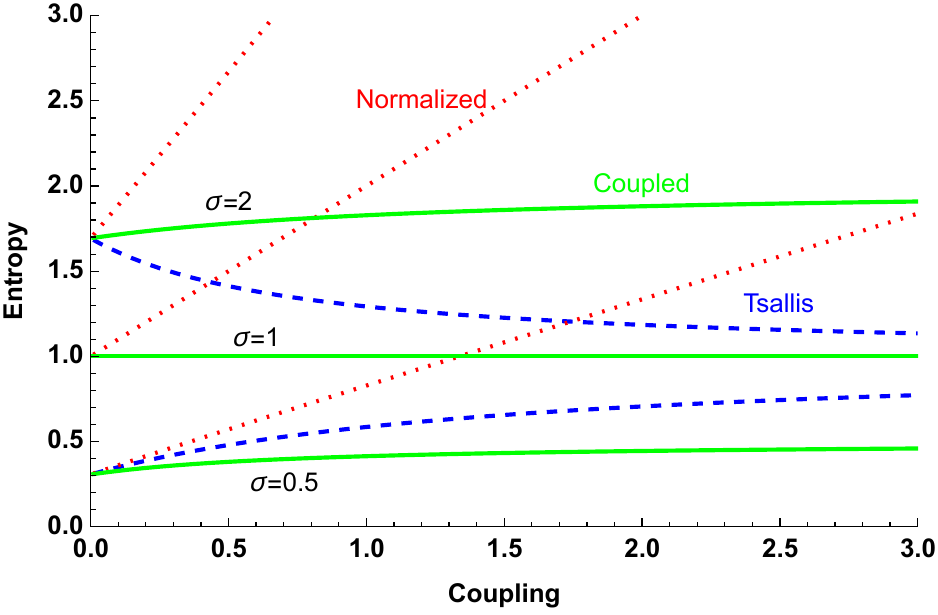}\label{fig:ents}
    \caption{Comparison of generalized entropies of the coupled exponential (generalized Pareto) distribution. The Tsallis entropy (blue) is too cold, converging to one as the coupling increases. The normalized Tsallis entropy (red) is too hot, increasing to infinity. Only the coupled entropy converges to the scale as the coupling increases.}
    
\end{figure}

Figure \ref{fig:ents} compares the coupled entropy with the Tsallis and normalized Tsallis entropy for the coupled exponential distributions. The visualization illustrate the rapid increase of the normalized Tsallis entropy due to the coupling multiplying the function of the scale. In contrast, the Tsallis entropy is shown to converge to one, thereby diminishing its ability to be a measure of uncertainty. 

This problem with the Tsallis entropy remained obscure due to the unnecessary complexity of the $q$-statistics framework, which is not grounded in physical measures. While $q$'s role in measuring the independent-equals was explained in the introduction, to the author's knowledge no prior investigators in the field of NSM explained this property. Instead, $q$ was often referred to as an index of the nonextensive and/or non-additive entropy \cite{plastino_nonextensive_2000, Wilk2000, tsallis_nonadditive_2009-1}, but of course it is not a measure of these properties. Further, $q$-statistics typically labels $\beta_q$ as the inverse width of the $q$-exponential distribution; however, as shown earlier this is not accurate. 

\subsection{Maximum Entropy Principle}
All of the theoretical foundations for nonextensive statistical mechanics will have to be revisited given the proposition that the coupled entropy function provides a unique fulfillment of the requirements for modeling the uncertainty of complex systems. Nevertheless, the structure of the NSM theoretical results will remain in tact. Here I provide a detailed proof of the maximum entropy principle. Regarding stability, the finite solution for the coupled entropy of the coupled exponential distribution is a strong indicator of its stability. For $\alpha\neq 1$, the Type II definition, which incorporates a root, is likely to have more stability than the Type III definition.

Khinchin's four axioms \cite{a_ya_khinchin_mathematical_1957} provide a uniqueness theorem for the BGS entropy. Modification or removal of the additivity axiom provides theoretical justification for generalizations of the BGS entropy \cite{zhang_khinchins_2023, thurner_entropy_2012,furuichi_uniqueness_2005,suyari_generalization_2004}. Thus a uniqueness proof using the following axioms would be a valuable contribution.
\begin{enumerate}
    \item Continuity: $H_{\alpha,\kappa}(\textbf{p)}$ is continuous with probability $\textbf{p}$.
    \item Maximality: $H_{\alpha,\kappa}(\textbf{p)}$ is maximized for $k\geq0$ by the uniform distribution $p_i=\sfrac{1}{N}$ for a discrete distribution with $N$ states. For $-\sfrac{1}{d}<\kappa<0$ the uniform distribution is minimal entropy.
    \item Coupled Additivity: $H_{\alpha,\kappa}(\textbf{p})$ violates the additivity axiom regarding the property of joint and conditional entropy, $H_0(X,Y)=H_0(X)+H_0(Y|X)$; however, a generalization of this axiom is anticipated given the property of coupled additivity for independent random variates $H_{\alpha,\kappa}(X,Y)=H_{\alpha,\kappa}(X)\oplus_{\alpha\kappa} H_(Y)$.
    \item Expandability: $H_{\alpha,\kappa}(\textbf{p})$ is defined to require $H_{\alpha,\kappa}(\textbf{p}_0)=H_{\alpha,\kappa}(\textbf{p})$ for $\textbf{p}_0={0,p_1,p_2,...}$; though this includes a discountinuity for  $-\sfrac{1}{d}<\kappa<0$.
\end{enumerate}

The connection between an entropy function and its maximizing distributions is established via the Lagrange functional. A foundational weakness in the development of the Tsallis entropy was the assumption that given a hypothesized entropy the maximizing distribution is derived and thus to be accepted without validation. As I established in the Background Section, in fact the definition of the shape-scale distributions was established correctly in the 1800s and the \textit{q}-exponentials have a weaker justification. Thus a stronger requirement for a generalized entropy is that a) the shape parameter and thus the nonlinearity or complexity defines the generalization, and b) that the maximizing distributions are the shape-scale distributions given the independent-equals constraints. Here I will derive the Type I Coupled Entropy with the coupled exponential distribution and the independent-equals mean constraint. The Type II and III will be based on the coupled Gaussian distribution; however, those derivations will be deferred given the merits of a more thorough investigation of the distinctions between these two forms. The extension to multivariate distributions should be straightforward but is also recommended for future research.

\begin{lemma}[Coupled Entropy Maximized by the Coupled Exponential Distribution]
    Given the coupled entropy Type I 
    \begin{equation} H_\kappa(p(x) = \int_X  
        P^{\left(1+\frac{\kappa}{1+d\kappa}\right)}(x)\ln_\kappa\left(p(x)\right)^{-\frac{1}{1+d\kappa}}=\frac{1}{\kappa}  \left(-1+\left(\int_X p^{1+\frac{\kappa}{1+d\kappa}}(x)\right)^{-1}\right)
    \end{equation} 
and  two constraints, the normalization and  the independent equals mean, $\mu_1^{(1+\frac{\kappa}{1+\kappa})}=\sigma$;
then, the maximum coupled entropy distribution is the coupled exponential, 
\begin{align}
p_\kappa^{\text{exp}}(x) &= \frac{1}{\sigma}\left(1+\kappa\frac{x-\mu}{\sigma}\right)^{-\frac{1+\kappa}{\kappa}}=\frac{1}{\sigma}\exp_\kappa^{-(1+\kappa)}\left(\frac{x-\mu}{\sigma}\right).
\end{align}
\end{lemma}
\begin{proof}
    Given the coupled entropy function and the two constraints the Lagrangian function is:
\begin{equation}
\mathcal{L} = H_\kappa(p) + \lambda_0 \left(1 - \int_0^\infty p(x)  dx \right) + \lambda_1 \left( \sigma -\int_0^\infty xP^{\left(1+\frac{\kappa}{1+\kappa}\right)} (x)dx\right),
\end{equation}
where $\lambda_0$ and $\lambda_1$ are the Lagrangian multiples for the normalization and independent equals mean, respectively. For maximization, the derivative must be zero, \(\delta \mathcal{L} / \delta p = 0\). The independent equals normalization is labeled $Z_P=\int_0^\infty p^{\frac{1+2\kappa}{1+\kappa}} (x)dx$ and the numerator is labeled $N_P=\int_0^\infty xp^{\frac{1+2\kappa}{1+\kappa}} (x)dx$.
The term by term derivatives of $\delta \mathcal{L} / \delta p $ are:
\begin{enumerate}
\item Entropy Derivative
            \begin{align}
                \frac{\delta H_\kappa(p(y))}{\delta p(y)} = 
                \frac{\delta}{\delta p(y)}\frac{1}{\kappa}\left(-1+Z_P^{-1}\right)
                 =-\frac{1+2\kappa}{\kappa(1+\kappa)}Z_P^{-2}p^\frac{\kappa}{1+\kappa}(y).
            \end{align}
\item The normalization derivative is $-\lambda_0$.
\item Independent equals derivative
            \begin{align}
                \frac{\delta}{\delta p(y)}  \lambda_1 \left( \sigma -\frac{N_P}{Z_P}\right) 
                &= - \lambda_1\frac{1+2\kappa}{1+\kappa}p^\frac{\kappa}{1+\kappa}(y)\frac{yZ_P-N_P}{Z_P^2}
            \end{align}
\end{enumerate}
Thus, the Lagrangian derivative is:
\begin{equation}\label{equ_Lder}
    \delta \frac{\mathcal{L}}{\delta p} =
    -\frac{1+2\kappa}{\kappa(1+\kappa)}p^\frac{\kappa}{1+\kappa}(y)Z_P^{-2}
    -\lambda_0
     - \lambda_1\frac{1+2\kappa}{1+\kappa}p^\frac{\kappa}{1+\kappa}(y)\frac{yZ_P-N_P}{Z_P^2}
     = 0.
\end{equation}
Solving for the probability, gives:
\begin{align}
    \lambda_0&=p^\frac{\kappa}{1+\kappa}(y)
    \left( -\frac{1+2\kappa}{\kappa(1+\kappa)}Z_P^{-2}
     - \lambda_1\frac{1+2\kappa}{1+\kappa}\frac{yZ_P-N_P}{Z_P^2}\right)\\
   p(y)&=\left(-\frac{1+2\kappa}{\kappa(1+\kappa)Z_P^{2}\lambda_0} 
   + \lambda_1\frac{(1+2\kappa)N_P}{(1+\kappa)Z_P^2\lambda_0}
   - \lambda_1\frac{1+2\kappa}{(1+\kappa)Z_P\lambda_0}y\right)^{-\frac{1+\kappa}{\kappa}}\\
\end{align}

which has the form 
\begin{equation}
    p(y)=\frac{1}{Z'}\left(1+\kappa\frac{y}{\sigma'}\right)^{-\frac{1+\kappa}{\kappa}}
    =\left(Z'^{\frac{\kappa}{1+\kappa}} + \kappa \frac{y}{\sigma'} Z'^{\frac{\kappa}{1+\kappa}} \right)^{-\frac{1+\kappa}{\kappa}}.
\end{equation} Applying the two constraints gives the equations:
\begin{enumerate}
    \item Normalization Constraint: $Z'=\sigma'$
    \item Independent Equals Constraint: $\sigma'=\sigma$
\end{enumerate}

Thus the solution is $p(y)=\frac{1}{\sigma}\left(1+\kappa\frac{y}{\sigma}\right)^{-\frac{1+\kappa}{\kappa}}$, confirming that the coupled exponential distribution maximizes $(\kappa\geq 0)$ or minimizes $-1<\kappa<0$ the coupled entropy.

The solutions for the Lagrangian multipliers are as follows. Note that raising the coupled exponential to  power $\frac{1+2\kappa}{1+\kappa}$ divides the shape and scale by $1+\kappa$: 
\begin{align}
    \frac{1+\kappa'}{\kappa'}&= \frac{1+\kappa}{\kappa}\frac{1+2\kappa}{1+\kappa}=2+\frac{1}{\kappa}\\
    \frac{\kappa'}{\kappa}&=\frac{1}{1+\kappa}=\frac{\sigma'}{\sigma}.
\end{align}
Then
\begin{align}
    Z_P&=\int_0^\infty p^{\frac{1+2\kappa}{1+\kappa}} (x)dx
    =\sigma^{-\frac{1+2\kappa}{1+\kappa}}\frac{\sigma}{1+\kappa}
    =\frac{1}{1+\kappa}\sigma^{-\frac{\kappa}{1+\kappa}}\\
    N_P&=\int_0^\infty xp^{\frac{1+2\kappa}{1+\kappa}} (x)dx
    =\sigma Z_p 
    =\frac{\sigma^\frac{1}{1+\kappa}}{1+\kappa}.
    \end{align}
The two equations are:
\begin{align}
    \frac{\kappa}{\sigma}\sigma^\frac{\kappa}{1+\kappa}&=- \lambda_1\frac{1+2\kappa}{(1+\kappa)Z_P\lambda_0}\\
    \sigma^\frac{\kappa}{1+\kappa}&=-\frac{1+2\kappa}{\kappa(1+\kappa)Z_P^{2}\lambda_0} 
   + \lambda_1\frac{(1+2\kappa)N_P}{(1+\kappa)Z_P^2\lambda_0}
\end{align}
Solving for $\lambda_0$ in terms of $\lambda_1$ and then for $\lambda_1$:
\begin{align}
     \lambda_0&=- \lambda_1\frac{1+2\kappa}{\kappa }\sigma\\
      \sigma^\frac{\kappa}{1+\kappa}&=\frac{1+\kappa}{\sigma^\frac{-2\kappa}{1+\kappa}\lambda_1 \sigma} 
   - \frac{\kappa\sigma}{\sigma^{-\frac{\kappa}{1+\kappa}}\sigma}\\
   \lambda_1(1+\kappa)\sigma^\frac{\kappa}{1+\kappa}
   &=(1+\kappa)\sigma^{-\frac{1-\kappa}{1+\kappa}}\\
   \lambda_1&=\sigma^{-\frac{1}{1+\kappa}};\ \lambda_0=-\frac{1+2\kappa}{\kappa}\sigma^\frac{\kappa}{1+\kappa}
\end{align}
\end{proof}

\section{Application to Physical and Informational Systems}\label{Sec_Appl}
\subsection{Generalization of the Boltzmann-Gibbs Distribution}
The generalized thermodynamics defined by NSM will need to be reexamined in light of the coupled entropy providing a more accurate computation of the non-additive entropy. A preliminary step for this is the properties of the generalized Boltzmann-Gibbs distribution. The most significant issue is the definition of the generalized temperature.  NSM has defined a generalized temperature based on $T_q=\frac{1}{k_B\beta_q}$, where $k_B$ is the Boltzmann constant. However, from the continuous GPD or CED it was established that $\frac{1}{\beta_q}$ for an infinite system would be $\frac{\sigma}{1+\kappa}$, and thus, this definition of the generalized temperature has an inverse dependence on the coupling. This creates the non-physical assertion that scale and shape of the distribution defining temperature can both go to infinity, while a supposed generalized temperature remains finite. Instead, a generalized temperature needs to separate out the dependence on the shape or nonlinearity, which necessitates replacing the $q$-statistic formulation with the coupled exponentials, which are in turn are consistent with the definitions established by Pareto. 

The generalize the Boltzmann-Gibbs distribution based on the coupled exponential distribution is:

\begin{definition}[Coupled Boltzmann-Gibbs distribution]
\begin{align}
    p_i&=\frac{1}{Z_\kappa}\exp_\kappa^{-(1+\kappa)}(\beta E_i)=\exp_\kappa^{-(1+\kappa)}\left(\ln_\kappa Z_\kappa^\frac{1}{1+\kappa}\oplus_\kappa\beta E_i\right)\\
    Z_\kappa&=\sum_{i=1}^W \exp_\kappa^{-(1+\kappa)}(\beta E_i) \text{\ is the partition function}\\
    \beta&=\frac{1}{k_BT} 
\end{align}
\end{definition}
The coupled entropy of the coupled Boltzmann-Gibbs distribution is:
\begin{align}
    S_\kappa(\textbf{p})&=\sum_{i=1}^W P_i^{(1+\frac{\kappa}{1+\kappa})}\ln_\kappa p_i^{-\frac{1}{1+\kappa}}\\
    &=\sum_{i=1}^W P_i^{(1+\frac{\kappa}{1+\kappa})}\left(\ln_\kappa Z_\kappa^\frac{1}{1+\kappa}\oplus_\kappa\beta E_i\right)\\
    &=\ln_\kappa Z_\kappa^\frac{1}{1+\kappa}\oplus_\kappa\beta U_\kappa.
\end{align}
The generalized internal energy of the system, $U_\kappa$ is defined by the first moment of the independent-equals distribution.
\begin{align}
    &U_\kappa=\langle E \rangle_\kappa=\sum_{i=1}^W P_i^{(1+\frac{\kappa}{1+\kappa})}E_i.
\end{align}
In the limit as $W\rightarrow \infty$, $\langle E \rangle_\kappa\rightarrow\frac{1}{\beta}$, given the continuous distribution result that $\int_0^\infty 
     \frac{x}{\sigma}\left(1+\kappa\frac{x}{\sigma}\right)^{-\frac{1+2\kappa}{1+\kappa}}=\sigma$.

An important research agenda will be to extend this preliminary step toward a thorough examination of the nonextensive statistical mechanics utilizing the coupled entropy.

\subsection{Application to Machine Learning}

One of the foundational methods of machine learning is variational inference (VI)\cite{blei_variational_2017}, in which a statistical model is learned and from which complex samples can be generated. VI seeks to minimize the divergence from a latent model $q(z)$ and the actual posterior probability given the dataset $p(z|x)$. Direct inference would require knowledge of $p(x)$ which is typically quite complex. Instead a variety of optimization methods have been developed that make learning $q(z)$ viable and enable generation of new samples via $p(x|z)$. An example is the variational autoencoder \cite{kingma_introduction_2019} in which an encoder network compresses the data distribution $p(x)$ into a posterior model $q(z|x)$ and the inverse of the network is used to decode samples from $q(z|x)$ into generated samples of $\hat{x}$. The optimization function for a VAE is the negative Evidence Lower Bound (-ELBO) or equivalently the Free Energy. The free energy has two components, the latent distribution divergence and the reconstruction negative log-likelihood. The divergence between the posterior and prior latent distribution, $D(q_\phi(z|x)||q_\theta(z))$, provides regularization toward a simple model. $\phi$ represents the parameters of the encoder and $\theta$ represents the parameters of the decoder. The negative log-likelihood measures the similarity of generated sample (or equivalently the cross-entropy between the generated and original samples), $\mathbb{E}_{q_\phi(z|x)}\left[\ln \left(p_\theta(x|z)\right)^{-1}\right]$.

A successful model must be flexible enough to enable modeling of complex datasets and simple enough to facilitate efficient computation. The coupled exponential family and its associated generalization of information theory contributes to improved VI models in two ways. First, the use of heavy-tailed (or compact-support) models enables the learning of more robust (decisive) models. Secondly, the coupled entropy functions enable learning that emphasizes (or de-emphasizes) samples in the tails of the distribution. The first design of the coupled VAE \cite{cao_coupled_2022}, which is similar to designs utilizing the Tsallis entropy, had difficulties maintaining stability of the learning process. As the coupling increased, the cost for low-likelihood generated samples increases rapidly. Eventually, the learning does not converge, thus these designs were limited to $\kappa<0.5$.

Recently, the independent equals distribution was used to draw from $Q_\kappa^{\left(1+\frac{\alpha\kappa}{1+d\kappa}\right)}(z|x)$ rather than $q(z|x)$ (the encoder/decoder symbols are dropped for simplicity) \cite{nelson_variational_2025}. If the latent distribution is a coupled Gaussian, the coupled probability divides both the scale and the shape by $1+\alpha\kappa$. By drawing from a lower-entropy distribution while using a cost-function that has higher penalties, there is a balance between robust sampling and robust system learning, respectively. The coupled free entropy function in this new design is 
\begin{equation}
\mathcal{F}_\kappa = \frac{1}{2}\mathbb{E}_{Q_\kappa^{\left(1+\frac{2\kappa}{1+d\kappa}\right)}(z|x)}\left[\ln_\kappa q(z)^{-\frac{2\kappa}{1+d\kappa}}+\ln_\kappa p(x|z)^{-\frac{2\kappa}{1+d\kappa}}\right].
\end{equation}

\section{Conclusion}\label{Sec_Con}
By grounding the nonextensive statistical mechanics framework in the fundamental property of complex systems, namely the nonlinearity, evidence is provided that that the coupled entropy provides a uniquely accurate generalization of statistical mechanics. Nonlinear sources of uncertainty, called the nonlinear statistical coupling, are equal to the shape of the generalized Pareto distribution and the inverse of the degree of freedom of the Student's t distribution. Examples include but are not limited to superstatistic fluctuations and multiplicative noise. Given the centrality of nonlinearity in defining the complexity of a system, the coupling can be considered a measure of complexity. 

The coupled entropy of the generalized Pareto distribution demonstrates the Goldilocks-like precision in which the BGS entropy is recalibrated to measure the uncertainty from the linear source while minimizing the influence of the nonlinear source. BGS entropy for the GPD is $1+\ln\sigma + \kappa$, while the coupled entropy is $1+\ln_\frac{\kappa}{1+\kappa}\sigma$. The uniqueness of this solution suggests that a more fundamental justification for the coupled entropy will be attainable. For instance, by specify the coupled sum with a nonlinear magnitude of $\alpha\kappa$ for the Khinchin expandablility axiom is expected to define its uniqueness. Complimenting the non-additivity, the nonextensivity is derived to be equal to the informational relative risk aversion $\frac{\alpha\kappa}{1+d\kappa}$.

Tsallis proposed a generalization of statistical mechanics using $q$-statistics and derived a function now known as the Tsallis entropy. Theoretical concerns led to an alternative known as the normalized Tsallis entropy but this proved to be unstable. This report clarifies that the normalized Tsallis entropy for the GPD includes a multiplicative term that contributes to its rapid growth with the coupling.  Overlooked is the fact that the Tsallis entropy has a term with inverse dependence on the statistical coupling, and it therefore converges to a constant. As such, as the complexity increases the Tsallis entropy fails to be a metric.  The problem with the Tsallis entropy arises from the fact that the parameters of the $q$-distributions do not properly define the shape and scale and therefore do not represent relevant physical properties. 

The coupled entropy framework retains the role of the escort distributions, identified as the distribution of independent equals; however, $q=1+\frac{\alpha\kappa}{1+d\kappa}$ is a secondary property that can not substitute for identifying the nonlinear statistical coupling. The coupled entropy has already facilitated advances in the training of robust variational inference algorithms. Since the optimizing function, the coupled free energy, utilizes the independent equals distribution, latent distributions with extreme heavy-tails can be trained reliably, since the training draws from distributions guaranteed to have a finite variance. The variational Lagrangian maximization for the coupled entropy was proved, motivating further theoretical research. A preliminary application of the coupled entropy and coupled free energy for the generalization of the Boltzmann-Gibbs distribution was reviewed. Further research is anticipated to greatly improve the accuracy with which non-additive and nonextensive principles can be applied to complex systems. The critical role nonlinearity plays in quantum, biological, socioeconomic, intelligence and other systems makes all the more relevant that generalizations of statistical mechanics be grounded in that nonlinearity. 

\section{Acknowledgements}
The author wishes to thank Igor Oliveira and Amenah Al-Najarif regarding applications of the coupled entropy and independent equals. DeepSeek and Mathematica were used to assist with the proofs and graphics. Wikipedia was used as a general reference, though the citations are to primary sources.

\printbibliography

\appendices
\section{Proof of Multiplicative Noise Process}\label{App_Noise}
\begin{proof}[Lemma 1 Proof: Multiplicative Process with Coupled Gaussian Limit]
The Fokker-Plank equation determines the stationary density \(p_s(x)\) in which the first derivative of the drift (mean) must balance the second derivative of the diffusion (local variance):  
\[
0 = -\partial_x [J(x) p_s] + \partial_x^2 [D(x) p_s].  
\]  
Substitute for \(J(x)\) and \(D(x)\) based on the relationships, $J(x)=f(X_t)$, $D(x)=\frac{1}{2}(A^2 + M^2 g^2(x))$, and $f(x)=-\tau g(x)g'(x)=-V'(x)$:  
\[
0 = \partial_x \left[ \tau g g' p_s \right] + \frac{1}{2} \partial_x^2 \left[ (A^2 + M^2 g^2) p_s \right].  
\]
One integration with zero flux boundary conditions gives:  
\[
\tau g g' p_s + \frac{1}{2} \partial_x \left[ (A^2 + M^2 g^2) p_s \right] = 0.  
\]
Rearranging terms forms the first-order ODE for \(p_s\):  
\[
\frac{\partial_x p_s}{p_s} = -2 \left( \frac{\tau g g' + \frac{M^2}{2} g g'}{A^2 + M^2 g^2} \right).  
\]
Using \(u = g^2(x)\) and \(du=2gg'\) the integral is:  
\begin{align*}
    \ln(p_s(x)) &\propto -\frac{2\tau + M^2}{2M^2}\ln{\left(A^2 + M^2 g^2(x)\right)}\\
    p_s(x)&\propto \left( A^2 + M^2 g^2(x) \right)^{-\frac{2\tau + M^2}{2M^2}}.  
\end{align*}
Equating the stationary density with the coupled Gaussian form determines the relationships:
\[
\kappa = \frac{M^2}{2\tau},\  \sigma^2 = \frac{A^2}{2\tau}.
\]
\end{proof}

\section{Proof of Entropies for Coupled Exponential}\label{App_Exp}

The entropy of the coupled exponential (generalized Pareto) distribution is equal to $1+\kappa+\ln \sigma$. A generalized entropy for the nonextensive statistical mechanics frameworks needs to eliminate the linear dependence  on the coupling (complexity) and generalize the logarithm function. The coupled entropy fulfills this requirement, while the Tsallis entropy is a) has an inverse dependence on the coupling and b) subtracts an inverse function of the scale. While the normalized Tsallis entropy corrects the relationship with the generalized logarithm of the scale, its linear and nonlinear dependence on the coupling causes it to go to infinity with the coupling.

\begin{proof}[Lemma 2 Proof: Entropies of the Coupled Exponential Distribution]
    The proof utilizes the first moment of the Independent-Equals.
    \begin{align*}
    H_\kappa\left(\text{Exp}_\kappa(\sigma)\right) &= \int_0^\infty  P^{(1+\frac{\kappa}{1+\kappa})}(x)
        \ln\left(\frac{1}{\sigma}\exp_\kappa^{-(1+\kappa)}\left(\frac{x}{\sigma}\right)\right)^{-\frac{1}{1+\kappa}}dx\\
    &=\int_0^\infty  P^{(1+\frac{\kappa}{1+\kappa})}(x)
        \left(\frac{x}{\sigma}\oplus_\kappa \ln_\kappa \sigma^\frac{1}{1+\kappa}\right)dx\\
        &=1+(1+\kappa)\ln_\kappa \sigma^\frac{1}{1+\kappa}\\
        &=1+\frac{1+\kappa}{\kappa}\left(\sigma^\frac{\kappa}{1+\kappa}-1\right)\\
         &=1+\ln_\frac{\kappa}{1+\kappa} \sigma
\end{align*}
    The normalized Tsallis entropy multiplies the coupled entropy by $1+\kappa$, this its measure of uncertainty for the coupled exponential distribution is 
     \begin{align*}
    H_\kappa^{NT}\left(\text{Exp}_\kappa(\sigma)\right) &= 1+\kappa +(1+\kappa)\ln_\frac{\kappa}{1+\kappa} \sigma
\end{align*}
The  Tsallis entropy multiplies the NTE by the normalization $\int_0^\infty  p^{(1+\frac{\kappa}{1+\kappa})}(x)dx$, thus its measure of uncertainty for the coupled exponential distribution is given by the following. The derivation makes use of the fact that raising the coupled exponential to the power $1+\frac{\kappa}{1+\kappa}$ converts the distribution into one with an effective coupling of $\frac{\kappa}{1+\kappa}$ and an effective scale of $\frac{\sigma}{1+\kappa}$.
\begin{align*}
     \int_0^\infty  p^{(1+\frac{\kappa}{1+\kappa})}(x)dx&=\int_0^\infty  \sigma^{-\frac{1+2\kappa}{1+\kappa}}\exp_\kappa^{-(1+\kappa)\frac{1+2\kappa}{1+\kappa}}\left(\frac{x}{\sigma}\right)dx\\
     &=\sigma^{-\frac{1+2\kappa}{1+\kappa}}\int_0^\infty  \exp_\frac{\kappa}{1+\kappa}^{-1}\left(\frac{1+\kappa}{\sigma}x\right)dx\\
     &=\sigma^{-\frac{1+2\kappa}{1+\kappa}}\frac{\sigma}{1+\kappa}=\frac{\sigma^{-\frac{\kappa}{1+\kappa}}}{1+\kappa}\\
    H_\kappa^{T}\left(\text{Exp}_\kappa(\sigma)\right) 
    &= \left(\frac{\sigma^{-\frac{\kappa}{1+\kappa}}}{1+\kappa}\right)\left(1+\kappa +(1+\kappa)\ln_\frac{\kappa}{1+\kappa} \sigma\right)\\
    &=\sigma^{-\frac{\kappa}{1+\kappa}}\left(1 +\frac{1+\kappa}{\kappa}\left(\sigma^{\frac{\kappa}{1+\kappa}}-1\right)\right)\\
    &=\sigma^{-\frac{\kappa}{1+\kappa}} +\frac{1}{\kappa}\left(1-\sigma^{-\frac{\kappa}{1+\kappa}}\right)+1-\sigma^{-\frac{\kappa}{1+\kappa}}\\
    &=1-\frac{1}{1+\kappa}\ln_\frac{\kappa}{1+\kappa} \sigma^{-1}\\
\end{align*}
\end{proof}
\end{document}